\documentclass[letterpaper, 10pt, conference]{ieeeconf}

\usepackage{amsmath,amssymb,amsfonts}
\usepackage{graphicx}
\usepackage{booktabs}
\usepackage{multirow}
\usepackage{xcolor}
\usepackage[hidelinks]{hyperref}
\usepackage{cite}
\usepackage[font=scriptsize,labelfont=bf,skip=2pt]{caption}

\newenvironment{IEEEkeywords}{\noindent\small\bfseries\itshape Index Terms---}{\par}

\newcommand{\id}{\textsc{id}}
\newcommand{\ood}{\textsc{ood}}
\newcommand{\bg}{\textsc{bg}}
\newcommand{\zvec}{\mathbf{z}}
\newcommand{\lvec}{\boldsymbol{\ell}}
\newcommand{\R}{\mathbb{R}}
\newcommand{\best}[1]{\textbf{#1}}      
\newcommand{\bestnf}[1]{\underline{#1}} 

\newif\ifreadfixes
\readfixestrue

\newif\ifrelworkfixes
\relworkfixestrue

\newif\ifopfixes
\opfixestrue

\begin{document}

\title{Three-Way Open-Set Detection for Robust Autonomous Navigation}
\author{Spyridon Loukovitis, Vasileios Karampinis, and Athanasios Voulodimos\\
 National Technical University of Athens \\
{\tt\small el20120@mail.ntua.gr, vkarampinis@ails.ece.ntua.gr, thanosv@mail.ntua.gr}}

\markboth{IEEE Robotics and Automation Letters. Preprint draft.}{}

\maketitle

\begin{abstract}
Autonomous navigation in complex scenes requires reliable perception across scenarios that the model did not encounter during its training. Along its route, an autonomous framework encounters objects it was trained to recognize, obstacles it has never seen, and background structures that resemble objects. Each of the three must be handled differently. To tackle this, existing open-set and out-of-distribution detectors discard low-confidence detections with an objectness threshold and only then test the rest for novelty. By forcing a single threshold like this they introduce a trade-off where a low threshold adds background clutter to the detected objects, while a high one may discard needed novel objects. We instead formulate open-set detection as a three-way classification of each detection into known object, unknown object, or background, computed post hoc from the outputs of a pretrained detector. We develop methods for domain generalization and for domain adaptation, evaluated across different detector families and benchmarks up to a combined semantic and covariate domain shift. To test the framework in a navigation context, we conduct simulations parameterized by the measured detection performance. The results show that the three-way decision yields safer and more efficient missions than binary alternatives.
\end{abstract}


\section{Introduction}\label{sec:intro}
Autonomous vehicles deployed for inspection, delivery, environmental monitoring, and search and rescue operate in cluttered, partially known environments. In these settings, perception is the primary safeguard for planning and manuvering around external objects.

More specifically, a detector operating in such scenes encounters objects of classes absent from training, as well as background structures that resemble objects. Safe navigation should treat each of the three differently. Known objects are the ones that can be handled more easily and confidently while unknown objects must be avoided with caution and background must be suppressed so that planning stays tractable. Unknown objects and background must also be kept apart, treating real obstacles as background means ignoring them and risks collision, while raising background as obstacles complicates the navigation plan and causes unnecessary maneuvers.

Typical detectors, however, operate on a closed set: they assign every detection to a known class and offer no way to flag a new one. Open-set and out-of-distribution (OOD) detection methods relax this limitation but they work in two separate steps. First, an objectness threshold removes weak detections. Then, a second test decides whether the remaining ones are known or novel. However, the threshold is set without accounting for how the novelty test will behave. A low threshold lets background clutter through, creating false obstacles that slow down the planner. A high threshold removes faint novel objects before they can be flagged, leading to missed obstacles.

We address this by formulating open-set detection as a three-way classification of each detection to known object (\id), unknown object (\ood), or background (\bg), computed post hoc from the outputs of a trained detector and applicable to any architecture. The decision combines two complementary signals already provided by the detector. Class-level confidence separates strong object hypotheses from diffuse background, while the distance between the detection's feature embedding and the known-class distribution tells known objects apart from novel ones. We study this decision under two data regimes. In domain generalization the methods see no unknown-class data at all, whereas domain adaptation grants a small budget of labeled unknown detections. To assess how the resulting decisions hold up across settings, we evaluate them on three detector families and three benchmarks whose distribution shift ranges from purely semantic novelty to combined semantic and covariate shift.\footnote{All benchmarks derive from public datasets \cite{lin2014coco,varma2019idd,aotdataset,arsenos2024corruptions}. The post-hoc heads, evaluation protocol, and simulation harness will be released upon acceptance.} Finally, closed-loop navigation simulations, parameterized by the measured detection performance, connect these metrics to mission outcomes.

In summary, this work makes the following contributions.
\begin{enumerate}
  \item A three-way per-detection decision with an explicit asymmetric cost, computed post hoc from two feature densities and requiring no unknown-class data (Secs.~\ref{sec:problem}--\ref{sec:method}).
  \item An ablation across three detector families and three benchmarks of increasing shift, showing that the decision transfers across architectures and outperforms single-confidence thresholding, with exceptions traced to embedding geometry (Secs.~\ref{sec:results}, \ref{sec:analysis}).
  \item A budget study showing that in-domain no outlier data is needed, while under combined shift roughly one hundred labeled detections suffice (Sec.~\ref{sec:results:budget}).
  \item Closed-loop navigation simulations showing that binary perception remains unsafe or inefficient at every threshold, whereas the three-way decision reaches a balanced regime (Sec.~\ref{sec:sim}).
\end{enumerate}

\section{Related Work}\label{sec:related}
\textbf{Open-set recognition and OOD detection.} A closed-set classifier is forced to assign every input to one of its training classes, so inputs from new classes are silently mislabeled. Open-set recognition addresses this by adding a reject option \cite{bendale2016openmax}, which raises the practical question of how to decide when to reject. OOD detection answers it by scoring how unusual an input is. The simplest score reads the classifier's own confidence by finding the maximum softmax probability \cite{hendrycks2017msp}. Since this confidence is often misleading on unfamiliar inputs, later works replaced it with the better-calibrated free energy of the logits \cite{liu2020energy}. A different route avoids the classifier output altogether and instead models the distribution of intermediate features, flagging inputs whose features are far from it \cite{lee2018mahalanobis}. Later, DDU \cite{mukhoti2023ddu} showed that a simple class-conditional Gaussian fit suffices for this purpose, a recipe we build on. All of these scores are computed without outlier data, when such data is available, outlier exposure sharpens them further \cite{hendrycks2019oe}. All, however, reduce novelty to a single scalar with a single threshold, making it impossible to distinguish novel objects from background.

\textbf{Open-set detection.} In object detection, the same question must be answered once per detection rather than once per image. Early work showed that detectors systematically confuse unknown objects with background \cite{dhamija2020elephant}. To mediate this, studies in robotic vision utilized uncertainty-based rejection of open-set errors through dropout sampling \cite{miller2018dropout}. Within this setting, GMM-Det \cite{miller2021gmmdet} fitted Gaussian mixtures to detector features to flag open-set errors, producing one open-set score per detection and introduced the evaluation protocol we adopt in this paper. SAFE \cite{wilson2023safe} follows the same post-hoc route with sensitivity-aware backbone features, but it reduces the outcome to a binary in/out flag. A separate family of methods takes the opposite approach and modifies the detector's training itself, shaping its feature space so that novel objects become easier to separate. This is done by synthesizing virtual outliers \cite{du2022vos}, expanding low-density latent regions \cite{han2022opendet}, or learning hyperspherical representations \cite{du2022siren}. These methods improve novelty separation at the source, but the price is that the detector must be retrained. Among them, UnSniffer \cite{liang2023unsniffer} comes closest to our target, since it explicitly separates unknown objects from background through a generalized-object confidence, yet it learns that separation during detector training, which ties it to each detector release.

\textbf{Open-world and open-vocabulary detection.} Open-world detection (OWOD) also flags unknown objects, but aims to incrementally learn them through an oracle-labeling and retraining loop \cite{joseph2021owod,zohar2023prob}, because it pools all unknowns into one bucket, scored by U-Recall and A-OSE, its gains arrive between sessions rather than per frame. Open-vocabulary detectors \cite{liu2023grounding,cheng2024yoloworld}, on the other hand, localize categories supplied as text and can serve as OOD scorers when prompted with the known-class names and thresholded on text similarity \cite{ming2022mcm}. However, they require the novel category to be nameable in advance and carry web-scale vision--language pretraining.

In contrast to all of the above, our work makes the unknown-versus-background distinction itself the decision target. We classify each detection three ways, post hoc, so the method transfers across architectures without retraining, and in its generalization regime it requires no unknown-class data. To our knowledge, no prior work evaluates this decision under an asymmetric cost or ties it to navigation outcomes.

\section{Problem Formulation}\label{sec:problem}
Let a single detection be summarized by its feature embedding $\zvec \in \R^{D}$ and its class-score vector $\lvec \in \R^{K}$ over the $K$ known classes. We seek a per-detection decision rule
\begin{equation}
  g:\ (\zvec,\lvec)\ \longmapsto\ y \in \{\,\id,\ \ood,\ \bg\,\},
\end{equation}
where $\id$ denotes a correctly localized known object, $\ood$ a localized object of a class never seen in training and $\bg$ a detection corresponding to no real object of interest. 

The decision is governed by an asymmetric cost $\mathcal{C}(\hat y \mid y)$:
\begin{equation}
  \mathcal{C}(\bg \mid \ood)\ \gg\ \mathcal{C}(\ood \mid \bg),
\end{equation}
since failing to flag a real novel obstacle risks collision while raising a false obstacle merely perturbs the plan. Because the two errors carry such different costs, a practical system must let the operator choose how conservative the decision is, that is, expose a tunable operating point. We set that operating point by demanding a target recall of true novel objects, and every method is evaluated at this common point (Sec.~\ref{sec:method:op}). In addition, we require the decision rule to be \emph{post-hoc}, meaning it operates on the outputs of the detector without modifying the detector itself. As a result, it refits in seconds when the domain changes, and the closed-set performance stays intact.

\section{Method}\label{sec:method}
We propose four methods that implement the three-way decision rule $g$ of Sec.~\ref{sec:problem}, labeled M0--M3 in the tables and ordered by how much information each one uses. Before detailing each method, we first define the two confidence families they draw on.

\subsection{Discriminative confidence}\label{sec:method:disc}
Let $p_k$ be the detector's posterior over known classes. We summarize how sharply it concentrates on a single class by
\begin{align}
  s_{\max} &= \max_{k} p_k, &
  H(\lvec) &= -\sum_{k} p_k \log p_k, \\
  E(\lvec) &= -\log \textstyle\sum_{k} e^{\ell_k}, &
  m(\lvec) &= \ell_{(1)} - \ell_{(2)},
\end{align}
where $\ell_{(1)},\ell_{(2)}$ denote the top two logits. Alongside these four quantities we also keep the detection score and the maximum raw logit $\ell_{(1)}$, which the detector already outputs, for six discriminative signals in total. A confident detection has large $s_{\max}$, small entropy $H$, low energy $E$, and a large margin $m$, an indecisive detection, characteristic of background, has the opposite. Each signal is oriented so that larger values indicate greater confidence (energy is scored as $-E$), placing all of them on a common $\id{>}\ood{>}\bg$ axis. This family answers whether the classifier commits to a known class, not whether the object actually is known.

\subsection{Generative confidence}\label{sec:method:gen}
Independently, we ask how typical the embedding $\zvec$ is of the features the model produces for genuine known objects. Following the deterministic-uncertainty recipe
\cite{mukhoti2023ddu,lee2018mahalanobis}, we fit one full-covariance Gaussian per known class $k$ to the embeddings of training-set detections that match known objects and score
\begin{equation}
  s_{\id}(\zvec) \;=\; \log \sum_{k=1}^{K} \mathcal{N}\!\big(\zvec;\ \mu_k, \Sigma_k\big).
\end{equation}
For numerical stability, we add a diagonal jitter shared across classes, set to the smallest order of magnitude that makes every class covariance positive definite, this alone conditions even the $D{=}1024$ Faster R-CNN covariances, so no dimensionality reduction is needed. Since detections on the training images that match no annotation are by definition background, we fit a single full-covariance Gaussian to their embeddings at no labeling cost, yielding the background typicality
\begin{equation}
  s_{\bg}(\zvec) \;=\; \log \mathcal{N}\!\big(\zvec;\ \mu_{\bg}, \Sigma_{\bg}\big).
\end{equation}
Both densities are estimated entirely from the detector's original training images, with no unknown-class labels.

\subsection{Complementarity of the two families}
The two families complement each other because they fail on different cases. The discriminative axis separates confident objects from diffuse background but is blind to novelty, since a novel object often triggers a confident posterior peaked at a visually similar known class. The generative axis separates typical from atypical features but cannot detect classification indecision. Each axis therefore covers only part of the three-way partition, and any binary known/unknown score ends up merging $\ood$ with either $\bg$ or $\id$ (Sec.~\ref{sec:results:decision}).

\subsection{Decision heads}\label{sec:method:m0}
\textit{Single-confidence partition (M0).} The simplest approach relies on a single confidence score. It places two thresholds $t_{lo}<t_{hi}$ on the detection score and assigns $\bg$ below $t_{lo}$, $\ood$ between them, and $\id$ above $t_{hi}$, on the premise that known objects score high, novel objects in between, and clutter low \cite{miller2021gmmdet}. Both thresholds are grid-searched to maximize macro-F1 on the training split. Since this split carries only known-object and background labels, the objective never sees an \ood\ example at fit time. As a result, it can not keep a middle band open and collapses the two thresholds onto each other ($t_{lo}{\approx}t_{hi}$), a behavior visible under M0's fitted rule in Table~\ref{tab:confusion}. M0 therefore consumes no unknown-class data and serves as the strongest single-axis baseline. Given this collapse of its fitted rule, its operating-point results rely instead on the method-agnostic novelty-recall sweep of Sec.~\ref{sec:method:op}, while its pairwise ranking is computed on the same confidence axis under the $\id>\ood>\bg$ ordering.

\textit{Open-space density rejection (M1).}\label{sec:method:m1} The pair $(s_{\id}, s_{\bg})$ supports the three-way decision without unknown-class data through open-space rejection. The reasoning is that a detection typical of neither known objects nor background lies in a region the model never modelled, so it can be declared novel. Since the two log-densities live on incomparable scales, we first standardize each one against its own fixed reference population, using known-object detections for $s_{\id}$ and background detections for $s_{\bg}$. This yields the z-scores $z_{\id}$ and $z_{\bg}$, from which we predict
\begin{equation}\label{eq:m1}
  \hat y \;=\; \arg\max\big(z_{\id},\, \tau_{\ood},\, z_{\bg}\big),
\end{equation}
whose three arguments correspond to \id, \ood, and \bg. Under this rule, a detection is declared novel exactly when both z-scores fall below the scalar threshold $\tau_{\ood}$. To set this threshold without any unknown-class data, we compute $\max(z_{\id}, z_{\bg})$ for every known-object and background detection in the training split and take its $q$-th percentile as $\tau_{\ood}$. This choice guarantees that at most a fraction $q$ of these detections can be flagged as novel, so $q$ acts as a budget on the false-novelty rate. We use $q{=}20\%$ throughout, and the operating-point sweep of Sec.~\ref{sec:method:op} covers the sensitivity to this choice. In total, M1 requires only the trained detector, its training images, and this single false-novelty budget.

\textit{Learned fusion with outlier supervision (M2, M3).}\label{sec:method:fusion}
When outlier data is available, we fuse the two confidence families with a small three-class MLP. Its input is the eight-dimensional vector formed by the six discriminative signals of Sec.~\ref{sec:method:disc} together with the two density scores. Raw embeddings are deliberately excluded, since they tend to overfit at small outlier budgets, a choice Sec.~\ref{sec:analysis:recover} revisits. The two variants differ only in the source of their outlier supervision. \emph{M2} relies on synthetic outliers alone. It samples artificial \ood\ training rows, one third of the real count, a ratio we did not tune, from outside the Mahalanobis shell that contains $90\%$ of the real $\id\,\cup\,\bg$ signal vectors. \emph{M3} instead draws real unknown-object detections from a disjoint image pool (Sec.~\ref{sec:setup:protocol}), capped at a budget $n \in \{10, 25, 50, 100, 250, 500, \text{full}\}$. Comparing M1 with M3($n$) then shows how much performance the no-unknown-data sacrifices, and how many labeled detections are needed to recover it.

\subsection{Spectral normalization of the backbone}\label{sec:method:sn}
The generative signal is trustworthy only if the feature map is smooth, so that off-distribution inputs land in low-density regions. Following DDU \cite{mukhoti2023ddu}, we apply soft spectral normalization \cite{miyato2018sn} to all backbone convolutions. Each detector is trained with and without SN from the same initial weights. The two differ by less than $0.005$ in closed-set AP, so any open-set difference reflects feature smoothness (Sec.~\ref{sec:results:sn}).

\subsection{Operating point}\label{sec:method:op}
Reflecting the asymmetric cost of Sec.~\ref{sec:problem}, all methods are compared at a common safety-driven operating point. Every head outputs three per-class scores and predicts the class with the highest one. To control how conservative this decision is, we add a constant offset $\tau$ to the \ood\ score before the comparison. Raising $\tau$ biases the decision toward \ood, so we sweep $\tau$ and keep the smallest value at which novelty recall reaches $80\%$, following the fixed-TPR practice of OOD evaluation \cite{liang2018odin,lee2018mahalanobis,liu2020energy}. The threshold-free macro-AUROC (Sec.~\ref{sec:setup:metrics}) ensures that ordering claims do not depend on this target. Note that $\tau$ differs from the fit-time budget $q$ of Sec.~\ref{sec:method:m1}, which limits false-novelty rate on known detections, whereas $\tau$ controls how many true novel objects are caught. This calibration is method-agnostic, since every method receives the same sweep, so performance differences at the operating point reflect only the quality of the underlying scores.

\section{Experimental Setup}\label{sec:setup}
\subsection{Benchmarks}\label{sec:setup:data}
We build three open-set benchmarks by designating a subset of each dataset's categories as known and the rest as unknown. The three benchmarks are ordered by increasing distribution shift between training and open-set evaluation. \emph{COCO-OS} is the GMM-Det split of COCO \cite{miller2021gmmdet,lin2014coco} with $50$ known and $30$ unknown classes. Since it is evaluated in-domain, its novelty is purely semantic. \emph{IDD} \cite{varma2019idd} is a driving benchmark with $12$ known classes and $3$ unknown ones, namely autorickshaw, animal, and vehicle-fallback, which are largely absent from common pretraining corpora. Finally, \emph{AOT-C\,$\to$\,RF} is the hardest case and the aerial scenario motivating this work. Here, detectors are trained to recognize airplanes and helicopters in corrupted Airborne Object Tracking imagery \cite{aotdataset,arsenos2024corruptions}, but are evaluated on the separately captured, low-light Real Flights recordings \cite{arsenos2024corruptions}, where the drone plays the role of the unknown class. Evaluation therefore differs from training both in semantics, through the novel class, and in imaging conditions, through the new low-light capture domain. This combined shift is the most challenging setting for the feature densities.

\subsection{Detectors}\label{sec:setup:detectors}
We evaluate five detector variants drawn from three architecture families, each trained per benchmark on the known classes. The first two families are Faster R-CNN \cite{ren2015faster} and RT-DETR \cite{zhao2024rtdetr}, which we train in two configurations, with and without a spectrally normalized backbone (Sec.~\ref{sec:method:sn}). The fifth variant is YOLOv8-L \cite{yolov8}. As the per-detection embedding $\zvec$, we use each family's natural representation. For Faster R-CNN this is the box-head fully connected feature ($D{=}1024$), for RT-DETR the final decoder query ($D{=}256$), and for YOLOv8 the penultimate convolutional feature of the classification branch at the detection's anchor ($D{=}256$). In all cases, no detector is modified or retrained for the open-set task.

\subsection{Protocol}\label{sec:setup:protocol}
Following GMM-Det \cite{miller2021gmmdet}, each detector is run once over each split and only detections with score $\ge 0.2$ are kept, a floor applied identically during fitting and evaluation. Each kept detection is then assigned a ground-truth label by IoU matching against the annotations. A detection overlapping a known-class box at IoU $\ge 0.5$ is labeled $\id$, one overlapping an unknown-class box at the same threshold is labeled $\ood$, and one below IoU $0.01$ is labeled $\bg$. Detections falling in the ambiguous band between these limits are discarded. Fitting and evaluation use strictly separate data. The Gaussian densities and the decision heads are fitted on the detector's own training split, while the open-set test split is divided evenly, image by image, into two disjoint sets. The first serves as an \ood\ pool from which M3 draws its budgeted outliers, and the second is the held-out evaluation set used by every method. On the evaluation side, the $\id$ population merges correctly classified and misclassified known-object detections, since dropping the latter inflates ID-vs-OOD AUROC by up to $+0.056$. Note that only objects the detector finds can be classified, and averaged over detectors the detection rate at the $0.2$ floor is $0.27$/$0.49$/$0.34$ for unknown classes versus $0.73$/$0.67$/$0.86$ for known ones. The M2/M3 fusion head is an MLP ($64$/$32$ hidden units, $400$ epochs, standardized inputs, no class re-weighting), trained in seconds on CPU.
\subsection{Metrics}\label{sec:setup:metrics}
We report three metrics. The threshold-free \emph{macro pairwise AUROC}, the mean of the three one-vs-one AUROCs ($\id$-vs-$\ood$, $\id$-vs-$\bg$, $\ood$-vs-$\bg$), measures ranking quality. Decision quality is evaluated at the $80\%$-novelty-recall operating point through macro-F1. \emph{Open-set mAP} measures known-class mAP after removing detections labeled $\ood$ or $\bg$, verifying that filtering preserves closed-set utility (Sec.~\ref{sec:results:map}). Bootstrap $95\%$ CIs yield margins of $\pm 0.006$--$0.014$ for AUROC and $\pm 0.015$--$0.024$ for F1. Unless noted otherwise, all discussed differences exceed these margins.

\section{Results}\label{sec:results}
The ablation is organized around the three questions raised in Sec.~\ref{sec:intro}, namely whether the three-way decision is achievable post hoc, whether it transfers across detector architectures, and what outlier supervision is worth.

\begin{table*}[t]
\centering
\footnotesize
\caption{Every cell shows two metrics for one detector on one benchmark: the \textbf{macro
pairwise AUROC}, measuring three-way ranking quality, followed by the \textbf{macro-F1} at the
$80\%$-novelty-recall operating point, measuring decision quality. M2 uses synthetic outliers
and M3 real ones, shown at $n{=}100$ and with the full pool.
Per metric, \textbf{bold} marks the best overall and \underline{underline} the best without real
\ood\ data.}
\label{tab:macro}\label{tab:opf1}
\scriptsize
\setlength{\tabcolsep}{3.2pt}
\renewcommand{\arraystretch}{0.88}
\begin{tabular}{llccccc}
\toprule
Benchmark & Detector & M0 (2-thr) & M1 (density) & M2 (synth) & M3 ($n{=}100$) & M3 (full) \\
\midrule
\multirow{5}{*}{COCO-OS}
 & FRCNN      & 0.685\,/\,0.367 & \bestnf{0.732}\,/\,\bestnf{\best{0.470}} & 0.722\,/\,0.356 & 0.722\,/\,0.360 & \best{0.749}\,/\,0.415 \\
 & FRCNN+SN   & 0.685\,/\,0.359 & \bestnf{\best{0.780}}\,/\,\bestnf{\best{0.516}} & 0.745\,/\,0.366 & 0.735\,/\,0.369 & 0.767\,/\,0.381 \\
 & RT-DETR    & 0.705\,/\,0.294 & \bestnf{0.819}\,/\,\bestnf{\best{0.483}} & 0.814\,/\,0.361 & 0.818\,/\,0.345 & \best{0.832}\,/\,0.396 \\
 & RT-DETR+SN & 0.690\,/\,0.296 & 0.796\,/\,\bestnf{0.446} & \bestnf{0.813}\,/\,0.386 & 0.816\,/\,0.346 & \best{0.829}\,/\,\best{0.471} \\
 & YOLOv8     & 0.734\,/\,0.305 & \bestnf{\best{0.803}}\,/\,\bestnf{\best{0.428}} & 0.770\,/\,0.320 & 0.771\,/\,0.320 & 0.786\,/\,0.325 \\
\midrule
\multirow{5}{*}{IDD}
 & FRCNN      & 0.717\,/\,0.279 & \bestnf{0.844}\,/\,\bestnf{0.445} & 0.832\,/\,0.319 & 0.852\,/\,0.317 & \best{0.892}\,/\,\best{0.487} \\
 & FRCNN+SN   & 0.725\,/\,0.293 & \bestnf{0.850}\,/\,\bestnf{\best{0.478}} & 0.835\,/\,0.299 & 0.862\,/\,0.347 & \best{0.899}\,/\,\best{0.478} \\
 & RT-DETR    & 0.710\,/\,0.214 & \bestnf{0.880}\,/\,\bestnf{0.522} & 0.854\,/\,0.306 & 0.871\,/\,0.330 & \best{0.919}\,/\,\best{0.572} \\
 & RT-DETR+SN & 0.710\,/\,0.215 & \bestnf{0.863}\,/\,\bestnf{0.482} & 0.852\,/\,0.285 & 0.870\,/\,0.327 & \best{0.920}\,/\,\best{0.623} \\
 & YOLOv8     & 0.743\,/\,0.212 & \bestnf{0.825}\,/\,\bestnf{0.362} & 0.810\,/\,0.244 & 0.854\,/\,0.322 & \best{0.895}\,/\,\best{0.398} \\
\midrule
\multirow{5}{*}{AOT-C$\to$RF}
 & FRCNN      & \bestnf{0.904}\,/\,0.343 & 0.865\,/\,\bestnf{0.498} & 0.850\,/\,0.330 & 0.936\,/\,0.537 & \best{0.954}\,/\,\best{0.733} \\
 & FRCNN+SN   & \bestnf{0.923}\,/\,0.346 & 0.901\,/\,\bestnf{0.519} & 0.907\,/\,0.347 & 0.952\,/\,0.508 & \best{0.955}\,/\,\best{0.739} \\
 & RT-DETR    & 0.859\,/\,0.382 & \bestnf{0.920}\,/\,\bestnf{0.524} & 0.849\,/\,0.327 & 0.957\,/\,0.699 & \best{0.974}\,/\,\best{0.857} \\
 & RT-DETR+SN & 0.873\,/\,0.358 & \bestnf{0.909}\,/\,\bestnf{0.468} & 0.827\,/\,0.317 & 0.964\,/\,0.743 & \best{0.972}\,/\,\best{0.821} \\
 & YOLOv8     & \bestnf{\best{0.832}}\,/\,0.415 & 0.667\,/\,\bestnf{\best{0.476}} & 0.727\,/\,0.408 & 0.820\,/\,0.441 & 0.808\,/\,0.454 \\
\bottomrule
\end{tabular}
\end{table*}

\begin{figure}[t]
\centering
\includegraphics[width=\columnwidth]{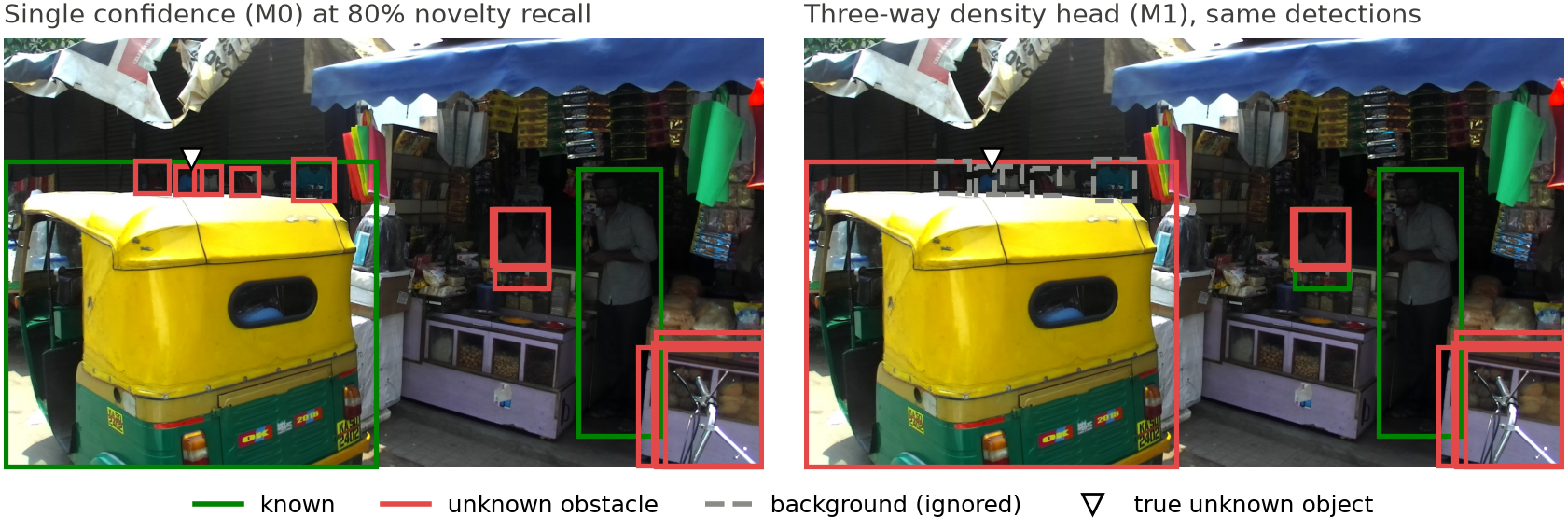}
\caption{Comparison of M0 (\textbf{left}) and M1 (\textbf{right}) on the same held-out IDD frame with identical RT-DETR detections. Green, red and gray-dashed boxes denote predicted known, unknown obstacle, and background and the white marker indicates the true unknown object. M0 mislabels the autorickshaw as approachable and raises background clutter as obstacles, whereas M1, using no unknown-class data, corrects both errors.}
\label{fig:qual}
\end{figure}

\subsection{Open-set recognition without unknown-class data}\label{sec:results:main}
Table~\ref{tab:macro} pairs every detector with every benchmark and reports two metrics for each such \emph{cell}: the threshold-free ranking quality and the operating-point decision quality. Fig.~\ref{fig:qual} illustrates the resulting decisions on an example frame. Starting with the two benchmarks without covariate shift, M1 improves macro pairwise AUROC over M0 in all ten cells, by $+0.05$--$0.11$ on COCO-OS and $+0.08$--$0.17$ on IDD. Moreover, these gains appear exactly where the complementarity analysis of Sec.~\ref{sec:method} predicts them. The single confidence axis cannot separate novel objects from background, reaching only $0.56$--$0.64$ $\ood$-vs-$\bg$ AUROC on IDD, whereas the density pair can, and it does so across two-stage, DETR-style, and one-stage detectors.

This ranking advantage carries over to the decision level. At the $80\%$-novelty-recall operating point, M1 exceeds M0 in all fifteen cells, with macro-F1 gaps of up to $+0.31$, reached by RT-DETR on IDD with $0.522$ versus $0.214$. Notably, M1 also holds up against the supervised heads. On COCO-OS, it remains best overall in four of five cells even against fusion with the full real-outlier pool, which shows that in-domain, unknown-class data adds little over open-space rejection.

Synthetic outliers do not help. M2 at best matches M1 on AUROC and is consistently worse at the operating point (e.g.\ F1 $0.356$ vs.\ $0.470$ on COCO-OS/FRCNN). Supervision pays off only when the outliers are real (Sec.~\ref{sec:results:budget}).

Under compound shift the picture changes. On AOT-C$\to$RF the ranking reverses for Faster R-CNN ($0.904$ vs.\ $0.865$) and YOLOv8 ($0.832$ vs.\ $0.667$), while RT-DETR retains the expected ordering. These reversals reproduce across SN variants and are traced to the detector's embedding geometry in Sec.~\ref{sec:analysis}.

\subsection{Ranking versus decision quality}
\label{sec:results:decision}

\begin{table}[t]
\centering
\footnotesize

\caption{The two non-SN reversed cells of Table~\ref{tab:macro} (AOT-C$\to$RF), as decisions
under each method's own fitted rule (M0: grid-searched thresholds, M1: native argmax). M0
maximizes F1 by nearly abandoning the \ood\ class.}
\label{tab:confusion}
\scriptsize
\setlength{\tabcolsep}{3.6pt}
\renewcommand{\arraystretch}{0.9}
\begin{tabular}{llccccc}
\toprule
Detector & Method & \id\ rec. & \ood\ rec. & \bg\ rec. & macro-F1 & acc. \\
\midrule
\multirow{2}{*}{YOLOv8} & M0 & 0.915 & 0.077 & 0.564 & 0.449 & 0.643 \\
                        & M1 & 0.323 & \textbf{0.875} & 0.270 & \textbf{0.451} & 0.483 \\
\midrule
\multirow{2}{*}{FRCNN}  & M0 & 0.975 & 0.036 & 0.836 & 0.525 & 0.706 \\
                        & M1 & 0.845 & \textbf{0.473} & 0.569 & \textbf{0.582} & 0.606 \\
\bottomrule
\end{tabular}
\end{table}

We examine the two non-SN reversed cells of Table~\ref{tab:macro} as three-way decisions, first at the shared operating point and then under each method's own fitted rule (Table~\ref{tab:confusion}). At the $80\%$-novelty-recall point, the threshold that achieves the mandated recall on a single confidence axis leaves no interval for background, so clutter floods the obstacle class and M0's macro-F1 drops to $0.343$--$0.415$ (AOT rows of Table~\ref{tab:macro}), below M1's $0.476$--$0.498$.

Table~\ref{tab:confusion} applies the second rule: M0 uses its grid-searched thresholds and M1 its native argmax, with no operating-point offset. Fitted without any \ood\ examples, the grid search collapses M0's middle band, and its novelty recall falls to $0.077$ for YOLOv8 and $0.036$ for Faster R-CNN. In effect, the baseline earns its competitive AUROC by collapsing into an $\id$/$\bg$ classifier.

M1, in contrast, keeps novelty recall at $0.875$ for YOLOv8 and $0.473$ for FRCNN at equal or better macro-F1 ($0.451$ vs.\ $0.449$ and $0.582$ vs.\ $0.525$), these F1 values differ from the $80\%$-recall ones above because no offset is applied here. M0's higher raw accuracy merely reflects the majority class. Together, these results confirm that a single confidence axis can rank well while being structurally unable to decide three ways.
\subsection{Published baselines}\label{sec:results:baselines}

\begin{table}[t]
\centering
\footnotesize
\caption{Published baselines under our protocol. Every cell shows macro AUROC followed by
operating-point F1, averaged over the five detectors. The binary scores use M0's two-threshold
rule, while the open-vocabulary YOLO-World runs standalone, prompted with the known classes.
Covers only the ${\sim}20$ detections YOLO-World finds on AOT.}
\label{tab:baselines}
\setlength{\tabcolsep}{2.2pt}
\resizebox{\columnwidth}{!}{%
\begin{tabular}{lcccccc}
\toprule
 & \multicolumn{2}{c}{COCO-OS} & \multicolumn{2}{c}{IDD} & \multicolumn{2}{c}{AOT-C$\to$RF} \\
\cmidrule(lr){2-3}\cmidrule(lr){4-5}\cmidrule(lr){6-7}
Method & AUROC & F1 & AUROC & F1 & AUROC & F1 \\
\midrule
MSP \cite{hendrycks2017msp}            & 0.674 & 0.324 & 0.699 & 0.243 & 0.874 & 0.369 \\
Max logit \cite{hendrycks2022scaling}  & 0.638 & 0.343 & 0.677 & 0.278 & 0.875 & 0.394 \\
Energy \cite{liu2020energy}            & 0.625 & 0.308 & 0.670 & 0.255 & 0.880 & 0.385 \\
Density, binary \cite{lee2018mahalanobis,miller2021gmmdet} & 0.575 & 0.240 & 0.610 & 0.201 & 0.753 & 0.224 \\
YOLO-World \cite{cheng2024yoloworld}   & 0.704 & 0.316 & 0.693 & 0.288 & 0.815$^\dagger$ & 0.347$^\dagger$ \\
\midrule
M0 (two-threshold)                     & 0.700 & 0.324 & 0.721 & 0.243 & 0.878 & 0.369 \\
M1 (density, three-way)                & 0.786 & 0.469 & 0.852 & 0.458 & 0.852 & 0.497 \\
\bottomrule
\end{tabular}%
}
\end{table}

Table~\ref{tab:baselines} places published alternatives under the same protocol. The standard binary OOD scores (MSP, max logit, energy) all perform at or below M0 and rank well below M1 at the operating point, confirming that the limitation of Sec.~\ref{sec:results:decision} is a property of single-axis scoring in general. Using the same feature density as a binary axis in the style of Mahalanobis-distance detectors \cite{lee2018mahalanobis,miller2021gmmdet} yields the weakest baseline ($0.240$ F1 on COCO-OS vs.\ M1's $0.469$), so the gain comes from the three-way rule over two densities, not the density signal alone.

The open-vocabulary detector YOLO-World \cite{cheng2024yoloworld}, prompted with the known-class names and scored by text similarity \cite{ming2022mcm}, trails M1 on all three benchmarks despite its web-scale pretraining. Under the compound aerial shift it produces only $0.14$ detections per image against $0.6$--$2$ for the fine-tuned detectors, so its ranking covers only the few objects it finds.
\subsection{Value of real outlier supervision (M3)}\label{sec:results:budget}
\begin{table}[t]
\centering
\footnotesize
\caption{Macro pairwise AUROC on AOT-C$\to$RF as M3's real-outlier budget $n$ grows, next to M1, which uses no \ood\ data. Gains saturate by $n\approx50$--$100$, while YOLOv8 plateaus early at a lower level (Sec.~\ref{sec:analysis}).}
\label{tab:budget}
\scriptsize
\setlength{\tabcolsep}{2.2pt}
\renewcommand{\arraystretch}{0.9}
\begin{tabular}{lcccccccc}
\toprule
Detector & M1 & $n{=}10$ & 25 & 50 & 100 & 250 & 500 & full \\
\midrule
FRCNN      & 0.865 & 0.910 & 0.909 & 0.933 & 0.936 & 0.939 & 0.946 & 0.954 \\
FRCNN+SN   & 0.901 & 0.936 & 0.922 & 0.929 & 0.952 & 0.944 & 0.935 & 0.955 \\
RT-DETR    & 0.920 & 0.940 & 0.959 & 0.965 & 0.957 & 0.968 & 0.974 & 0.974 \\
RT-DETR+SN & 0.909 & 0.949 & 0.951 & 0.954 & 0.964 & 0.967 & 0.971 & 0.972 \\
YOLOv8     & 0.667 & 0.817 & 0.815 & 0.811 & 0.820 & 0.805 & 0.803 & 0.808 \\
\bottomrule
\end{tabular}
\end{table}

Table~\ref{tab:budget} answers the final question by sweeping the real-outlier budget under the compound shift. The effect of real supervision appears immediately, as even $n{=}10$ real unknown detections lift macro AUROC by $+0.02$--$+0.15$ over M1. Beyond $n\approx 50$--$100$ the curves flatten, reaching $0.95$--$0.97$ for the CNN and DETR families. The reason for this rapid gain is that the fused head learns how the shifted unknowns actually score, which repairs the calibration that the covariate shift broke. In contrast, the same supervision is largely ineffective in-domain. On COCO-OS the best M3 exceeds M1 by at most $+0.033$, falls behind it in two of the five cells of Table~\ref{tab:macro}, and loses to it at the operating point in four of five. Taken together, these results show that collecting target-domain outliers pays off only under domain shift.

\subsection{Open-set filtering and closed-set mAP}\label{sec:results:map}

\begin{table*}[t]
\centering
\footnotesize
\caption{Open-set mAP (known-class mAP after removing predicted \ood/\bg) vs.\ the frozen
detector's closed-set mAP, both on a $0$--$1$ scale. M3 uses the full outlier pool.}
\label{tab:map}
\scriptsize
\setlength{\tabcolsep}{3.4pt}
\renewcommand{\arraystretch}{0.88}
\begin{tabular}{l cccc c cccc c cccc}
\toprule
 & \multicolumn{4}{c}{COCO-OS} & & \multicolumn{4}{c}{IDD} & & \multicolumn{4}{c}{AOT-C$\to$RF} \\
Detector & closed & M0 & M1 & M3 & & closed & M0 & M1 & M3 & & closed & M0 & M1 & M3 \\
\midrule
FRCNN      & 0.291 & 0.281 & 0.198 & 0.276 & & 0.162 & 0.155 & 0.121 & 0.152 & & 0.323 & 0.321 & 0.290 & 0.324 \\
FRCNN+SN   & 0.307 & 0.294 & 0.226 & 0.297 & & 0.155 & 0.148 & 0.120 & 0.149 & & 0.327 & 0.327 & 0.318 & 0.327 \\
RT-DETR    & 0.403 & 0.386 & 0.347 & 0.397 & & 0.202 & 0.197 & 0.168 & 0.199 & & 0.374 & 0.369 & 0.273 & 0.380 \\
RT-DETR+SN & 0.410 & 0.401 & 0.370 & 0.390 & & 0.204 & 0.199 & 0.167 & 0.199 & & 0.348 & 0.344 & 0.232 & 0.364 \\
YOLOv8     & 0.311 & 0.292 & 0.230 & 0.309 & & 0.187 & 0.181 & 0.125 & 0.186 & & 0.380 & 0.367 & 0.154 & 0.373 \\
\bottomrule
\end{tabular}
\end{table*}

Open-set filtering must not destroy closed-set utility. To check this, we measure known-class mAP after removing detections labeled \ood\ or \bg\ and compare it with the detector's closed-set mAP (Table~\ref{tab:map}). M0 retains it almost perfectly, losing only $0.001$ to $0.019$ mAP, but this is simply because it rarely removes anything as \ood. M1, on the other hand, shows a measurable mAP reduction under its default rule. It loses $0.01$ to $0.12$ across most cells, and up to $0.226$ in the YOLOv8/AOT cell examined in Sec.~\ref{sec:analysis}. The reason is that faint known objects are also atypical under the densities, so filtering out atypical detections inevitably removes some of them. This reduction can be controlled, however, by adjusting the operating point of Sec.~\ref{sec:method:op}, which trades retained mAP against novelty recall. Moreover, real outlier data removes the trade-off almost entirely. With the full outlier pool, M3 restores mAP to within $0.02$ of the closed-set value, and on AOT-C$\to$RF it even matches or exceeds it.

\subsection{Backbone spectral normalization}\label{sec:results:sn}
SN leaves closed-set AP essentially unchanged ($\leq 0.005$). For Faster R-CNN it raises M1's AUROC by up to $+0.048$; for RT-DETR it lowers it slightly ($-0.011$ to $-0.023$) and leaves M3 unchanged. SN smooths CNN features toward the Gaussian shape the densities assume, whereas RT-DETR's queries are already near-Gaussian, so it helps exactly the detectors whose features are density-hostile.

\begin{figure*}[t]
\centering
\includegraphics[width=0.80\textwidth]{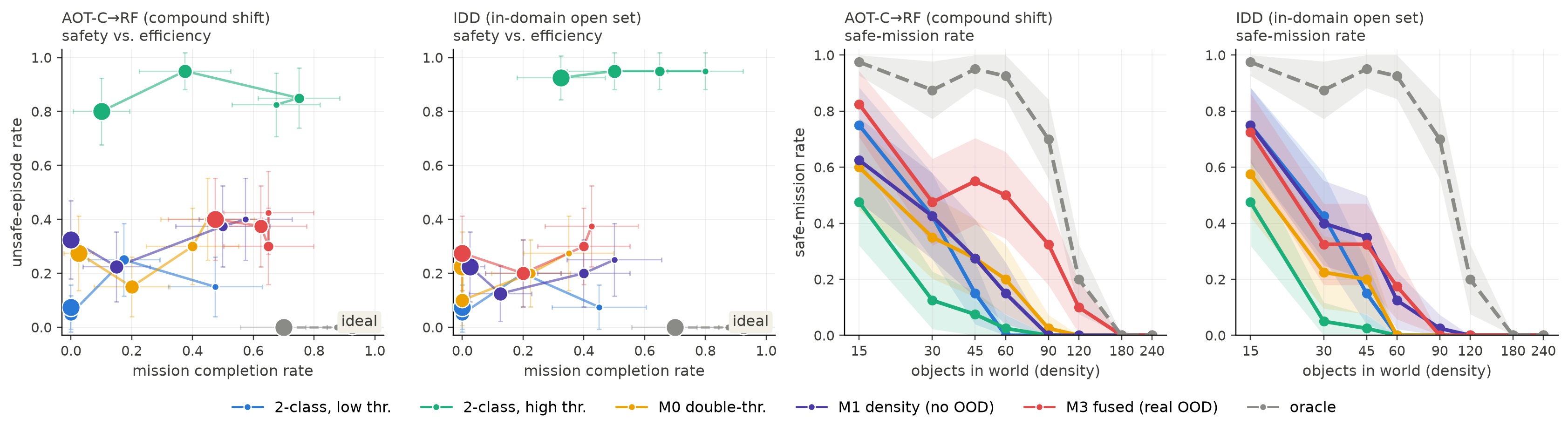}
\caption{Navigation simulations (RT-DETR confusion matrices, $95\%$ CIs over 40 seeds). \textbf{Left pair:} safety/efficiency plane over the $30\to90$ density sub-range (marker size grows with density, ideal corner is bottom-right). \textbf{Right pair:} safe-mission rate over the full $15\to240$ sweep, upper-bounded by the dashed oracle.}
\label{fig:sim}
\end{figure*}

\section{Analysis of Feature-Density Failures}\label{sec:analysis}
The three reversed cells all appear on AOT-C$\to$RF and involve only Faster R-CNN and YOLOv8, suggesting a feature-side cause. We isolate it with controlled experiments on YOLOv8 (the extreme case), using the other detectors as comparisons.

\subsection{Information content versus density access}\label{sec:analysis:geometry}

\begin{table}[t]
\centering
\footnotesize
\caption{Embedding geometry vs.\ density usefulness (AOT-C$\to$RF, $\id$-vs-$\ood$ AUROC). The
supervised logistic probe and the unsupervised density $s_{\id}$ read the same raw embeddings,
the probe--density gap predicts failure severity.}
\label{tab:geometry}
\scriptsize
\setlength{\tabcolsep}{2.4pt}
\renewcommand{\arraystretch}{0.9}
\begin{tabular}{lccccc}
\toprule
Detector & frac.\ $\le 0$ & exc.\ kurtosis & probe & density & gap \\
\midrule
RT-DETR & 0.51 & 0.1   & 0.98 & 0.865 & 0.11 \\
FRCNN   & 0.66 & 146   & 0.96 & 0.784 & 0.18 \\
YOLOv8  & 0.90 & 106   & 0.90 & 0.636 & 0.27 \\
\bottomrule
\end{tabular}
\end{table}

The first experiment asks whether the novelty information is missing from the embeddings themselves or merely inaccessible to the Gaussian density. To answer this, Table~\ref{tab:geometry} compares two readings of the same embedding vectors for each detector. The first is a supervised linear probe, which measures how much novelty information is linearly accessible in the embeddings. The second is the unsupervised Gaussian score $s_{\id}$, which measures how much of that information the density actually extracts. Since the probe needs a small labeled outlier sample, this comparison serves as a diagnostic rather than a deployable method.

The probe finds strong signal in every detector, with AUROC between $0.90$ and $0.98$, so the information is present in all embeddings. How much of it the density recovers, however, depends on how closely the embedding distribution resembles a Gaussian. RT-DETR's decoder queries are dense and nearly Gaussian in shape, with an excess kurtosis of just $0.1$, and accordingly the density loses only $0.11$ of the probe's AUROC. YOLOv8 sits at the other extreme. Its embeddings come from a post-activation convolutional feature, where $90\%$ of the values are at or below zero and the excess kurtosis reaches ${\sim}106$. In such a sparse code, class identity is carried by which few channels fire. The next linear layer and the probe can read this pattern, but an elliptical log-density cannot.

Across the three detectors, the probe--density gap grows from $0.11$ to $0.18$ to $0.27$, and this ordering matches the ranking reversal in Table~\ref{tab:macro}. Importantly, neither sparsity nor kurtosis alone predicts the failure, since Faster R-CNN's kurtosis exceeds YOLOv8's. The reliable diagnostic is therefore the gap itself, which can be measured from known-class data plus a small outlier sample.

Two further controls exclude alternative explanations for the failure. The first candidate explanation is background-sample scarcity, since YOLOv8 yields only 758 background detections to fit its 256-dimensional Gaussian. To test this, we subsample RT-DETR from its 16{,}910 background detections down to the same 758. Its $\ood$-vs-$\bg$ separation barely moves, dropping from $0.917$ to $0.909$, while YOLOv8 sits at $0.593$, so sample size is not the cause. The second candidate is poor covariance conditioning. However, even aggressive shrinkage of YOLOv8's background covariance recovers only $0.70$, which remains far below the $0.83$ that the raw confidence axis achieves.

\subsection{A controlled study of extraction points}\label{sec:analysis:extraction} If sparse post-activation features were the whole problem, extracting the embedding elsewhere in the same head should repair M1. To test this, we instrumented the YOLOv8 head at six per-detection extraction points, covering both classification convolutions and the penultimate box convolution, each taken before and after activation. We then re-ran the identical density pipeline on every alternative embedding.

The denser pre-activation features do carry more linearly readable novelty, with the probe reaching up to $0.961$ compared to $0.903$ for the default point, and they also repair the background-related axes. Despite this, the unsupervised Gaussian extracts less $\id$-vs-$\ood$ signal from every alternative, scoring $0.34$--$0.54$ against the default's $0.636$. The reason is that novel and known objects at these extraction points overlap in feature space at similar densities and they are separated only along a discriminative direction that no class-conditional Gaussian can represent.

This result yields a testable design rule. Feature densities need representations in which novel objects land in low-density regions, a property that DETR-style query embeddings satisfy but sparse one-stage head activations may not, regardless of the extraction point.
\subsection{Discriminative recovery within the outlier budget}\label{sec:analysis:recover} The probes above used the full evaluation pool, so they only bound the available information. To test whether it can also be recovered under the realistic M3 protocol, we train an $L_2$-regularized logistic head directly on the raw pre-activation classification features, using $n{=}100$ pooled outliers. This head reaches a held-out $\id$-vs-$\ood$ AUROC of $0.893 \pm 0.007$ over five subsampling seeds, rising to $0.942$ with the full pool. At the same budget, the scalar-fusion M3 reaches only $0.757$ and the confidence axis only $0.741$.

This result reveals two things. First, excluding raw embeddings from learned open-set heads out of fear of overfitting is unnecessary here, since regularization keeps the head stable even at a budget of $100$ detections. Second, the scalar-fusion design itself is what capped YOLOv8's M3 plateau, because that design sees the embeddings only through the broken density scores. The one-stage detector's novelty information is thus inaccessible generatively but recoverable discriminatively at the cost of a hundred labeled boxes.

\section{From Detection Metrics to Navigation Outcomes}\label{sec:sim}
To translate these metrics into mission outcomes, we built a navigation simulation in which each per-detection classification is sampled from the measured confusion matrix (RT-DETR on each benchmark), so the simulation inherits both the strengths and errors of each method.

\textbf{Setup.} A simulated drone (double integrator, $|a|{\le}2.5$\,m/s$^2$, $|v|{\le}4$\,m/s, $\Delta t{=}0.2$\,s) must cross an $80{\times}80{\times}24$\,m volume of equal parts \id, \ood, and \bg\ objects (placed uniformly, static). Ground-truth rules are to approach knowns to $3$\,m, hold $12$\,m standoff from unknowns, and treat background as traversable. Objects within $30$\,m are classified by sampling the confusion matrix. Two-class baselines use objectness thresholds $\theta{=}0.25$ and $\theta{=}0.75$, representing the two extremes available to a system without a background label. Labels persist per viewpoint, redrawn only after ${\ge}5$\,m or ${\ge}25^\circ$ of motion. An A* planner on a $2$\,m voxel grid (omnidirectional sensor, no occlusion) replans on any belief change, treating each object's clearance as unreachable. A two-class system must apply one class-agnostic $12$\,m standoff

We sweep object count from $15$ to $240$ at $40$ seeds each, scoring mission completion, unsafe episodes (collision or standoff-zone intrusion), safe-mission rate (both conditions met), and median completion time.

Figure~\ref{fig:sim} shows the results. The low-threshold agent completes $0\%$ of missions from $60$ objects on, while the high-threshold one is unsafe in $80$--$95\%$ of runs and produces the study's only collisions. The three-way heads escape this dilemma. In-domain, M1 already matches M3 at low unsafe rates and near-oracle finish times. Under compound shift only M3 closes the gap, reaching completion $0.62$ versus M1's $0.15$ at $60$ objects. M0 tracks the low-threshold corner because it cannot allocate a background class (Sec.~\ref{sec:results:decision}). In the broken-density cells of Sec.~\ref{sec:analysis} the three-way advantage disappears, so the simulation reflects perception quality rather than flattering the method.

\section{Limitations and Future Work}\label{sec:limitations}
The main limitation concerns the no-real-data regime under compound shift. As the M1/M3 contrast showed, M2's synthetic outliers are drawn in the fused scalar-signal space around the in-distribution data, so when the domain shifts, M2 tracks M1 rather than closing the gap toward M3. Shift-aware synthetic outliers, such as those from corruption or illumination augmentation, could teach the fused head this shifted geometry without requiring real unknowns.

We also do not compare against retraining-based open-set detectors such as VOS, OpenDet, SIREN, and OWOD, since these methods modify the detector itself and a fair comparison would require matched retraining. We leave this for future work.

\section{Conclusion}\label{sec:conclusion}
We extended binary open-set detection to the three-way decision that navigation needs, separating known objects, unknown obstacles, and background. The decision is computed post hoc on a trained detector by combining classifier confidence with feature typicality. When no unknown-class data is available, open-space rejection consistently improves both the three-way ranking and the operating-point decision over single-confidence thresholding, which cannot allocate a third class. Under domain shift, a small budget of real outlier detections repairs the density model and recovers closed-set-level detection quality. In the few cells in which the density head does fail, our analysis traces the failure to the embedding geometry and shows that the novelty signal remains recoverable by discriminative means. In simulations driven by the measured detection performance, binary perception remains unsafe or inefficient at every threshold, whereas the three-way decision reaches a balanced safety/efficiency regime (Fig.~\ref{fig:sim}). The natural next step is to fold the recovered discriminative signal into the budgeted head, together with shift-aware synthetic outliers.

\bibliographystyle{IEEEtran}
{\footnotesize \bibliography{refs}}

\end{document}


\title{Supplementary Material:\\ Three-Way Open-Set Detection for Robust Autonomous Navigation}
\author{}
\maketitle

\section{Real-outlier budget sweep}
Table~\ref{tab:budget} reports the full budget sweep summarized in the main paper: macro pairwise
AUROC of M3 trained with $n$ real unknown detections on AOT-C$\to$RF, against M1 (no \ood\ data).

\begin{table}[h]
\centering
\caption{Real-outlier budget sweep on AOT-C$\to$RF: macro pairwise AUROC of M3 trained with $n$
real unknown detections, against M1 (no \ood\ data). Gains saturate by $n\approx50$--$100$;
YOLOv8's plateau is the feature-geometry ceiling analyzed in the main paper.}
\label{tab:budget}
\setlength{\tabcolsep}{2.4pt}
\begin{tabular}{lcccccccc}
\toprule
Detector & M1 & $n{=}10$ & 25 & 50 & 100 & 250 & 500 & full \\
\midrule
FRCNN      & 0.865 & 0.910 & 0.909 & 0.933 & 0.936 & 0.939 & 0.946 & 0.954 \\
FRCNN+SN   & 0.901 & 0.936 & 0.922 & 0.929 & 0.952 & 0.944 & 0.935 & 0.955 \\
RT-DETR    & 0.920 & 0.940 & 0.959 & 0.965 & 0.957 & 0.968 & 0.974 & 0.974 \\
RT-DETR+SN & 0.909 & 0.949 & 0.951 & 0.954 & 0.964 & 0.967 & 0.971 & 0.972 \\
YOLOv8     & 0.667 & 0.817 & 0.815 & 0.811 & 0.820 & 0.805 & 0.803 & 0.808 \\
\bottomrule
\end{tabular}
\end{table}

\section{Open-set filtering and closed-set mAP}
Table~\ref{tab:map} reports the per-cell retention numbers summarized in the main paper.

\begin{table}[h]
\centering
\caption{Known-class mAP after open-set filtering (open-set mAP) vs.\ the frozen detector's
closed-set mAP. M0 preserves mAP by rarely predicting \ood; M1 pays a retention cost at its
default rule; M3 (full) recovers closed-set-level mAP together with the open-set gains reported in
the main paper.}
\label{tab:map}
\setlength{\tabcolsep}{4.5pt}
\begin{tabular}{llcccc}
\toprule
Benchmark & Detector & closed & M0 & M1 & M3 (full) \\
\midrule
\multirow{5}{*}{COCO-OS}
 & FRCNN      & 29.1 & 28.1 & 19.8 & 27.6 \\
 & FRCNN+SN   & 30.7 & 29.4 & 22.6 & 29.7 \\
 & RT-DETR    & 40.3 & 38.6 & 34.7 & 39.7 \\
 & RT-DETR+SN & 41.0 & 40.1 & 37.0 & 39.0 \\
 & YOLOv8     & 31.1 & 29.2 & 23.0 & 30.9 \\
\midrule
\multirow{5}{*}{IDD}
 & FRCNN      & 16.2 & 15.5 & 12.1 & 15.2 \\
 & FRCNN+SN   & 15.5 & 14.8 & 12.0 & 14.9 \\
 & RT-DETR    & 20.2 & 19.7 & 16.8 & 19.9 \\
 & RT-DETR+SN & 20.4 & 19.9 & 16.7 & 19.9 \\
 & YOLOv8     & 18.7 & 18.1 & 12.5 & 18.6 \\
\midrule
\multirow{5}{*}{AOT-C$\to$RF}
 & FRCNN      & 32.3 & 32.1 & 29.0 & 32.4 \\
 & FRCNN+SN   & 32.7 & 32.7 & 31.8 & 32.7 \\
 & RT-DETR    & 37.4 & 36.9 & 27.3 & 38.0 \\
 & RT-DETR+SN & 34.8 & 34.4 & 23.2 & 36.4 \\
 & YOLOv8     & 38.0 & 36.7 & 15.4 & 37.3 \\
\bottomrule
\end{tabular}
\end{table}

\section{Extraction-point study inside the YOLOv8 head}
If sparse post-activation features were the whole problem, extracting elsewhere in the same head
should repair the density head. We instrumented the frozen YOLOv8 head at six per-detection
extraction points---both convolutions of the classification branch and the penultimate convolution
of the box branch, each pre- and post-activation---and re-ran the identical density pipeline
(Table~\ref{tab:extraction}). The denser pre-activation features carry more linearly readable
novelty (probe up to $0.961$ vs $0.903$ for the default) and repair the background-related axes
(macro up to $0.708$ from $0.667$), but the unsupervised Gaussian extracts less $\id$-vs-$\ood$
signal from every alternative ($0.34$--$0.54$ vs $0.636$). In this cell the novel and known
objects (drones vs.\ aircraft, small and distant) occupy overlapping regions of feature space at
similar densities; what separates them is a discriminative direction, not a low-density region,
and no relocation of a class-conditional Gaussian can exploit a direction it cannot represent.

\begin{table}[h]
\centering
\caption{Six embedding extraction points inside the frozen YOLOv8 head (AOT-C$\to$RF), each
evaluated by the M1 density machinery and by a supervised linear probe. Denser (pre-activation)
features carry more novelty information, yet the unsupervised density extracts less: the
bottleneck is the density model, not the extraction point.}
\label{tab:extraction}
\setlength{\tabcolsep}{4.0pt}
\begin{tabular}{lcccc}
\toprule
Embedding (Detect head) & $D$ & \multicolumn{2}{c}{M1 density} & probe \\
\cmidrule(lr){3-4}
 & & \id-vs-\ood & macro & \id-vs-\ood \\
\midrule
cls conv 2, post-act (default) & 256 & \textbf{0.636} & 0.667 & 0.903 \\
cls conv 2, pre-act            & 256 & 0.526 & \textbf{0.708} & 0.945 \\
cls conv 1, post-act           & 256 & 0.539 & 0.705 & 0.954 \\
cls conv 1, pre-act            & 256 & 0.536 & 0.696 & \textbf{0.961} \\
box conv 2, post-act           & 64  & 0.370 & 0.598 & 0.877 \\
box conv 2, pre-act            & 64  & 0.336 & 0.592 & 0.886 \\
\midrule
\multicolumn{2}{l}{reference: M0 confidence axis} & 0.741 & 0.832 & -- \\
\bottomrule
\end{tabular}
\end{table}

\section{Full fitted-rule confusion matrices}
Table~\ref{tab:conffull} gives the complete row-normalized confusion matrices summarized in the
main paper's decision-level comparison.

\begin{table}[h]
\centering
\caption{Fitted-rule confusion matrices (true class $\times$ predicted) for the two non-SN
reversed cells on AOT-C$\to$RF.}
\label{tab:conffull}
\setlength{\tabcolsep}{4.2pt}
\begin{tabular}{llccc c c}
\toprule
 & & \multicolumn{3}{c}{predicted} & \ood\ recall & macro-F1 \\
\cmidrule(lr){3-5}
 & true & \id & \ood & \bg & & \\
\midrule
\multirow{6}{*}{\rotatebox{90}{YOLOv8}}
 & \multicolumn{6}{l}{\emph{M0 (double threshold)} \hfill acc.\ $0.643$} \\
 & \id  & 0.915 & 0.031 & 0.054 & \multirow{3}{*}{0.077} & \multirow{3}{*}{0.449} \\
 & \ood & 0.789 & 0.077 & 0.134 & & \\
 & \bg  & 0.294 & 0.143 & 0.564 & & \\
\cmidrule(lr){2-7}
 & \multicolumn{6}{l}{\emph{M1 (density-reject)} \hfill acc.\ $0.483$} \\
 & \id  & 0.323 & 0.670 & 0.007 & \multirow{3}{*}{\textbf{0.875}} & \multirow{3}{*}{\textbf{0.451}} \\
 & \ood & 0.104 & 0.875 & 0.020 & & \\
 & \bg  & 0.000 & 0.730 & 0.270 & & \\
\midrule
\multirow{6}{*}{\rotatebox{90}{FRCNN}}
 & \multicolumn{6}{l}{\emph{M0 (double threshold)} \hfill acc.\ $0.706$} \\
 & \id  & 0.975 & 0.004 & 0.021 & \multirow{3}{*}{0.036} & \multirow{3}{*}{0.525} \\
 & \ood & 0.789 & 0.036 & 0.175 & & \\
 & \bg  & 0.108 & 0.056 & 0.836 & & \\
\cmidrule(lr){2-7}
 & \multicolumn{6}{l}{\emph{M1 (density-reject)} \hfill acc.\ $0.606$} \\
 & \id  & 0.845 & 0.152 & 0.003 & \multirow{3}{*}{\textbf{0.473}} & \multirow{3}{*}{\textbf{0.582}} \\
 & \ood & 0.509 & 0.473 & 0.018 & & \\
 & \bg  & 0.037 & 0.394 & 0.569 & & \\
\bottomrule
\end{tabular}
\end{table}